\documentclass[letterpaper]{article}
\usepackage{aaai}
\usepackage{times}
\usepackage{helvet}
\usepackage{courier}
\usepackage{booktabs}
\usepackage{multirow}
\makeatletter
\newcommand{\@BIBLABEL}{\@emptybiblabel}
\newcommand{\@emptybiblabel}[1]{}
\makeatother
\usepackage[hidelinks]{hyperref}
\usepackage{scalerel}           
\usepackage{xparse}

\usepackage{graphicx}
\usepackage[toc,page]{appendix}

\NewDocumentCommand\emojione{}{\includegraphics{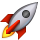}{}}

\begin{document}
%
\title{Turning Stocks into Memes: A Dataset for Understanding How Social Communities Can Drive Wall Street}
\author{Richard Alvarez, Paras Bhatt, Xingmeng Zhao, \and Anthony Rios\\
Department of Information Systems \& Cyber Security\\
University of Texas at San Antonio\\
San Antonio, TX 78249\\
}
\maketitle
\begin{abstract}
\begin{quote}
Who actually expresses an intent to buy GameStop shares on Reddit? What convinces people to buy stocks? Are people convinced to support a coordinated plan to adversely impact Wall Street investors? Existing literature on understanding intent has mainly relied on surveys and self reporting; however there are limitations to these methodologies. Hence, in this paper, we develop an annotated dataset of communications centered on the GameStop phenomenon to analyze the subscriber intentions behaviors within the r/WallStreetBets community to buy (or not buy) stocks. Likewise, we curate a dataset to better understand how intent interacts with a user's general support towards the coordinated actions of the community for GameStop. Overall, our dataset can provide insight to social scientists on the persuasive power to buy into social movements online by adopting common language and narrative.  \textbf{WARNING: This paper contains offensive language that commonly appears on Reddit's r/WallStreetBets subreddit.} 
\end{quote}
\end{abstract}

\section{Introduction}
\noindent There has been substantial research in persuasion and social engineering with a particular interest in how individuals can be convinced to behave (i.e., persuasion), buy a product or support some idea. For instance, \citeauthor{caldas2019persuasion} uses surveys to understand how the political framing of online discourse can impact ``buy-in'' to specific ideas. Likewise, in \citeauthor{wang2021social}, the authors evaluate how content (e.g., videos, text, etc.) and content creator characteristics affect the likelihood that people support or purchase a product. An interesting research direction is to understand the persuasive techniques used on social media platforms (e.g., Reddit) to get users to buy a product or support a specific social campaign. Before we can understand what drives persuasion, we must detect who intends to buy a product or at least supports the general ideology behind it. Hence, in this paper, we develop a new resource that can help researchers better understand purchase intentions and expressions of support on Reddit. 

Surveys and interviews have been used in prior research to understand persuasion, which requires knowing the user's intent to buy products or support campaigns. For example, research in online marketing focuses on measuring persuasiveness of an argument using self-reported surveys~\cite{caldas2019persuasion,{gerlach2019they}}, or by quantifying how responsive individuals are to tailored arguments in an online survey~\cite{axt2020psychological,ormond2019integrating,wang2021social}. While there have been recent advances in understanding persuasion and attitudes towards products and marketing campaigns~\cite{wang2021mitigating,pignot2020affective}), the studies are not entirely practical (e.g., using executive messaging for interviews~\cite{pignot2020affective}) or do not extract direct intentions to purchase~\cite{wang2021mitigating}.  Hence, the motivation for this paper is to provide social researchers with a dataset to extract social support for an online campaign as well as users intentions; thereby providing researchers with a new way to conduct quantitative studies about buying intentions and support, which can potentially be used for downstream research on persuasion and social manipulation.

\begin{figure}[t]
    \centering
    \includegraphics[width=0.5\linewidth]{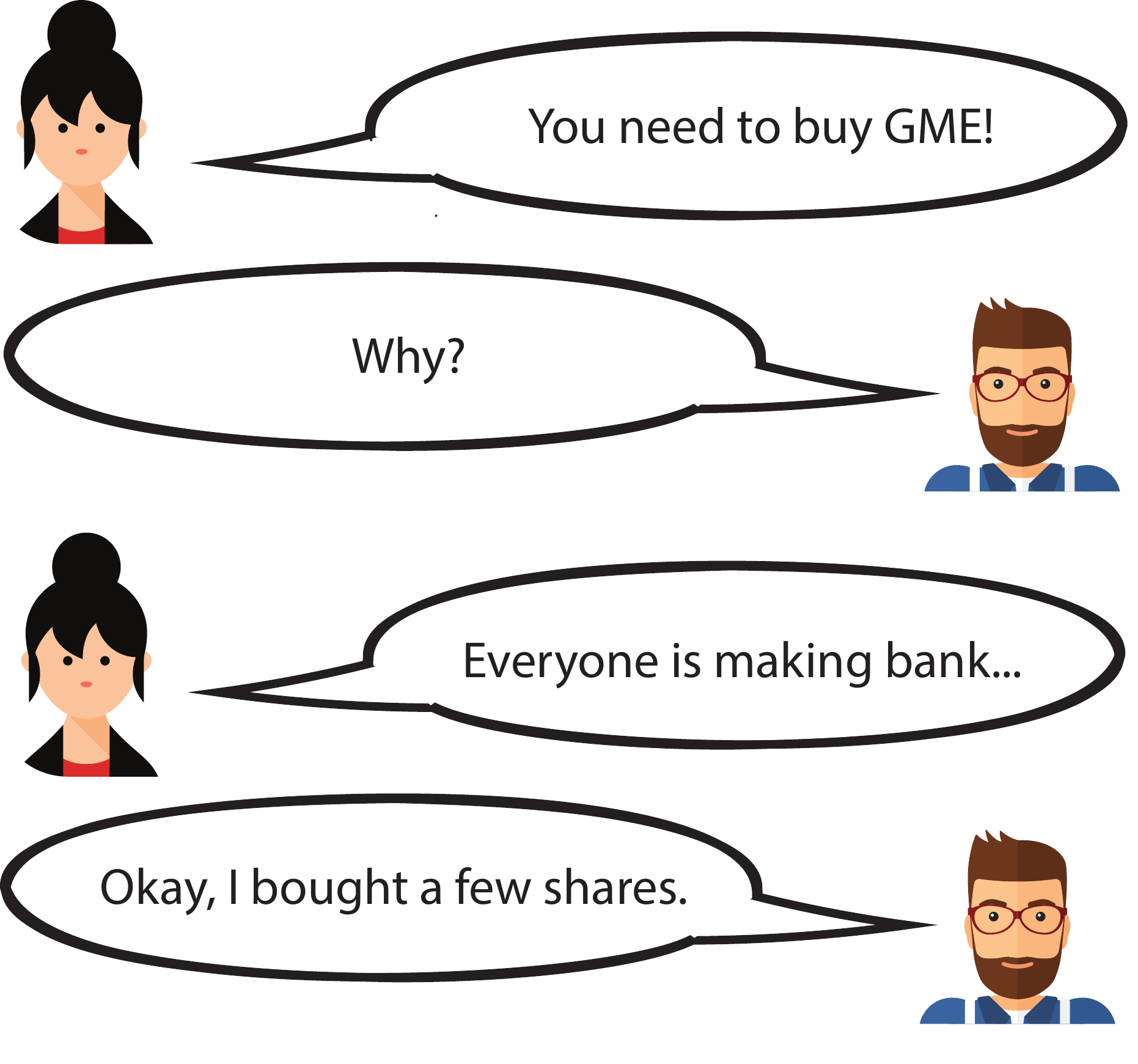}
    \caption{An example conversation on Reddit where a user is convinced to buy GME stock.}
    \label{fig:example}
    \vspace{-1em}
\end{figure}

Specifically, we introduce a new dataset that uses comments from the Reddit community r/WallStreetBets (WSB), which substantially influenced  GameStop~\footnote{We use GME throughout the paper to refer to the GameStop stock. GME is the ticker symbol for GameStop.} in early 2021. An example conversation that may appear on r/WallStreetBets is shown in Figure~\ref{fig:example}. We can see that the last comment directly mentions someone's positive intent to buy a product (GME shares). One specific use case for our dataset is to facilitate the development of machine learning models to extract these direct intentions. Our aim for this dataset is to help researchers study both direct intent (General \textit{Intent}) and the gradual evolution of persuasion by measurements of supporting metrics (General \textit{Support}) tied to social manipulation. For instance, in Figure 1 the user was persuaded to buy the GME shares. Is this a one-off item, or are there specific patterns that exist that cause someone to buy? Hence, an important use case of our dataset is to use a model developed to predict intent to identify all positive and negative purchase intentions which can be used as a dependent variable for various discourse analyses. 

Particularly, our dataset captures both purchase \textit{Intent} as well as varying degrees of \textit{Support} for GME and related campaigns. As previously stated, Intent is related to whether someone actually intends (or already did) purchase GME shares, while Support is focused on the general anti-Wall Street narrative.
Specifically, the GameStop narrative grew on Reddit into a David vs. Goliath-Esque narrative between the Redditors (representing ordinary investors) and big hedge funds (representing faceless corporations). Starting from December 2020, WSB saw an explosion of threads and comments centered on GameStop being a struggling videogame retail company. Rather than simply a stock recommendation, the stock experienced a short squeeze where according to the narrative, large faceless hedge funds were trying to kill the ailing corporation Gamestop by betting against its success. Whether users saw themselves as heroes, defending the company and main street from large scale investors and forcing the price up; the desire to ``stick it to the establishment''~\cite{businside} and exact revenge for the 2008 market collapse, or simply exhibiting the fear of missing out (referred to as FOMO on WallStreetBets) users were persuaded to purchase the stock. For this dataset, we collected data on developing events of GME, the favorite stock of WallStreetBets during the period.

What makes this circumstance particularly interesting from a persuasion perspective is that each individual who ends up purchasing GameStop stock is heavily incentivized to sell when it is high (prior to the event, the stock was worth only \$18 a share, while it peaked at more than \$480 per share). Yet, even after catching more than a tenfold increase in price (and therefore considerable profit), individuals were persuaded to hold onto the stock together. As long as the retail investor held, the more damage that could be done to the hedge funds and the greater the profit that could be extracted.
The factors that contribute to this decentralized unity amongst the members of these online communities is strong, causing individuals to ignore the financial incentives of selling out are interesting.
Therefore, an individual would have to be convinced to ``buy-in'' to the investment opportunity. We hope that our dataset to detect intent and support can be used to understand such phenomena.

Overall, this dataset and paper make several contributions to the existing literature. First, this dataset provides an easily accessible and annotated dataset of peer-to-peer conversations online between anonymous and semi-anonymous individuals. Such availability will assist researchers, particularly in replicability and transparency for their studies. Furthermore, the accessibility of the dataset will hopefully encourage further studies into social intent and online social manipulations. One advantage this study has over similar ones is the verification of intent. Individuals who express intent to engage in action may not always follow through. This dataset would be possible to track intent and eventual completion of the action (a transition not easily traceable in existing literature).
Second, we provide the results of several baseline models showcasing that the models trained on our data can provide accurate predictions for Intent and Support.
Third, we provide detailed future use-cases for our dataset to answer interesting social science-related questions.

\section{Related Work} 

This section describes the main areas of research related to this paper: Stocks and Financial NLP, Aspect-Based Sentiment Analysis (ABSA), and Stance Detection.

\subsection{Stocks and Financial NLP}
Research in forecasting stock market activity has been a mainstay of NLP-based studies that leverage content from the financial industry. For example, \citeauthor{das2007yahoo} use NLP to facilitate ``news‐based trading,'' wherein analysts seek to isolate financial news that affects stock prices and/or market activity. \citeauthor{seo2002text} used Natural Language Processing (NLP) data, processed with various combinations of feature extraction (e.g., Latent Semantic Analysis and a Naive Bayes classifier and a weighted‐majority voting ensemble, to analyze news articles, with the optimal combination yielding a 79\% accurate classification of articles that signaled an increase or decrease in stock prices. Similarly, \citeauthor{seo2002text} processed text from web‐based stock‐discussion bulletin boards, analyzed the output using the Naive-Bayes‐based classifier algorithm and multiple runs through a genetic algorithm, and generated significant (p$<$0.0001) excess returns.

In another study, \citeauthor{yildirim2018classification} classified financial news articles as ``hot'' (significant) and ``non-hot'' (non-significant) to study their impact as predictors on stock price forecasting. In time, multiple NLP‐based approaches were used to explore the predictive value of various international accounting and finance‐related text sources. \citeauthor{zhai2007combining} applied part-of-speech features and TF‐IDF-weighting, enhanced by Gaussian‐radial‐basis‐function‐kernel and polynomial‐kernel supervised SVM classifiers, to confirm correlations between the textual content of financial news articles and stock‐price trends in Australia. \citeauthor{lugmayr2013evaluation} proposed analyzing broker newsletters with a German‐based sentiment analysis SVM (LUGO Sentiment Indicator) to predict Deutscher Aktienindex German Stock Index (DAX 30) trading activity levels. \citeauthor{hagenau2013automated} employed bi‐normal separation‐based feature selection, enhanced by an SVM classifier, to predict stock‐price changes signaled by German financial news with 71.8\% precision. Argentine and Brazilian currency trends were successfully predicted by  \citeauthor{jin2013forex}, using their Forex‐Foreteller system, which employed topic clustering, sentiment analysis (based on the Loughran–McDonald, and AFINN sentiment‐analysis dictionaries) and regression analysis.

Finally, there has been a recent surge in research exploring the r/WallStreetBets community. For example, \citeauthor{buz2021should} evaluate whether people should take investment advice from the community, finding that many buy signals on Reddit can result in gains. \citeauthor{wang2021predicting} use data from r/WallStreetBets to predict stock movement. \citeauthor{mendoza2021sticking} qualitatively explore r/WallStreetBet's general sentiment towards the anti-establishment narrative.
Overall, our work expands on prior work by considering both the task of predicting whether someone intends to buy a particular stock as well as extracting their Support for the GME-related ``take down the establishment''  campaign for quantitative analysis. Our dataset allows for analyses of both financial interest and general political/social ideology. Moreover, our work is less focused on how the social behaviors affect the company and instead how the individual is persuaded to make the purchase on a micro level. Ultimately, our dataset will provide subsequent research on conversation-level and user-level persuasion and purchase-intent on Reddit.

\subsection{ABSA and Stance Detection}
There has been substantial work in understanding various incarnations of Support in the NLP community. Support can be thought of as some sort of valence (sentiment) towards a specific entity (ABSA) or simply whether someone agrees or disagrees with a specific topic (Stance Detection).  \citeauthor{demszky2019analyzing} apply ABSA methods to tweets about US mass shootings topic, where the topic was politically discussed from different viewpoints according to the locations of events with the contrasting use of the terms ``terrorist'' and ``crazy'', that contribute to polarization.  \citeauthor{chen2020aspect} show that most recent ABSA approaches rely on state-of-the-art supervised approaches combining complex layers of neural network models (e.g., transformers) to classify labels representing aspects from text elements with the standard sentiment (i.e., positive, negative). 

Stance detection refers to the task of classifying a piece of text as either being in support, opposition, or neutral towards a given target. The most well-known data for political stance detection is published by the SemEval 2016~\cite{mohammad2016semeval}. The paper describing the data set provides a high-level review of approaches to stance detection using Twitter data. The best user-submitted system was a neural classifier from \citeauthor{zarrella2016mitre} which utilized a pre-trained language model on a large amount of unlabeled data. An important contribution of this study was using pre-trained word embeddings from an auxiliary task where a language model was trained to predict a missing hashtag from a given tweet. Likewise, \citeauthor{wei2016pkudblab} show that convolutional neural networks also perform accurately for the task.

Contrary to prior work on ABSA and Stance detection, our work differs in one important aspect. Specifically, we label positive support towards and event/idea even if that idea is not explicitly mentioned within the text. For example, the comment ``I'm never going to sell my GME shares!!!'' would show positive support for. Traditional ABSA and methods may only calculate a sentiment score with respect to nouns in the sentence (e.g., GME Shares). We note that there has been some recent work on implicit sentiment using Connotation Frames~\cite{sap2017connotation}. However, the work is focused at verb understanding and does not perform classification over an entire comment.


\section{Methodology}
In this section we discuss the methodology involved in data collection, annotation, and evaluating annotation quality.

\subsection{Data Collection}

We collected data using the Python Pushshift.io API Wrapper (PSAW)~\cite{baumgartner2020pushshift} library to collect submissions and comments in the r/WallStreetBets community. Submissions refer to descriptive posts made on the discussion board of the community by its members. Each submission has a ‘title’ and a corresponding ‘body’ of text that represents the main idea discussed within the submission. We have captured these data items separately in our dataset. This is because compared to other social media communities (Twitter, Facebook, etc.) Reddit is more of a discussion-based forum where people can talk about anything.To interact with one another, redditors will join smaller subject based groups referred to as subreddits. 
We scraped the r/WallStreetBets subreddit for all threads during the period January 1, 2021, to March 1, 2021, that contained some mention of GameStop or GME. The collected data is comprised of two complementing datasets, the Reddit posts, which included the author, postdate, ID, posting category, number of comments, author cross-posts, whether the comment was pinned, the comment score, the post submission text, the post ID number, the post title, and the ratio of upvotes to downvotes. The second set tracks the accompanying post ID and lists the commenter ids, the time and date of the comment, the reputation score of the comment, and the comments themselves. Both datasets are linked through the Post ID number. Overall, we obtained a total of 71,075 submissions and 100,069 comments. From the entire dataset of comments, we randomly sampled 5,000 to annotate.

\subsection{Data Annotation}
For this study, we first took an exploratory approach to analyze the comments for the WSB community. The comments corpus was randomized utilizing a randomization python script. 
Initially, a pilot annotation was done with ten observations to establish relevant annotation rules. Comments were reviewed for two primary characteristics; the intent of the comment (Does the comment indicate an intention to purchase or has purchased GameStop during the craze period?) and level of support (Does the comment indicate support for the WSB community or support for the David vs. Goliath narrative?). These two annotation categories were chosen to capture both the level of community support as well as to identify if the support had a tangible impact. A second annotator was brought in to annotate another set of 50 observations using the pilot rule set for refinement. After multiple revisions, both parties would agree to the finalized annotation ruleset and guidelines. The final guidelines can be found in the Appendix.

\textbf{Intent Annotations.} The first annotation category, \textit{Intent}, refers to the general intent to purchase the stock. We wanted to capture all statements that suggest the individual plans on or has already purchased the GME investment for this category. The category is broken into five different annotations: Yes, Maybe, Informative, Unknown, and No. We describe each category below:
\begin{itemize}
\item \textit{Yes} indicates there is clear intent to purchase or has already purchased GME shares in the recent time-period. 

\item \textit{Maybe} indicates uncertainty that the individual has the stock, but the context hints of a possibility of purchase or already owning the stock. 

\item \textit{Informative} posts are meant to capture potential moderator or bot comments. These are meant to inform users without any personal opinions or biases visible.  No emotions, no sides taken, only sharing information.

\item \textit{Unknown} It is not clear one way or another the intent to purchase or currently owning the stock. This can serve as a catchall if unable to annotate to any other category, such as completely unrelated posts.
\item \textit{No} indicates a clear disinterest, or no intention to purchase the stock. The individual does not and will not purchase the stock. Alternatively, the individual could be betting against the stock, hoping to bring it down.
\end{itemize}

\textbf{Support Annotations.} The second annotation category, \textit{Suport}, is to measure the degree of buy-in the individual has with the current narrative. We defined the narrative as either support for the Us vs. them mentality (that is, support for GameStop because it hurts the institutions), support for the hype (that is, support the camaraderie, to be part of the moment, or to see it as a historical moment in the making), or alternatively sees GameStop as a legitimate investment. This category was broken into the following annotation classes:

\begin{itemize}
\item \textit{Yes} posts indicate clear support for the GME narrative. For GameStop as a company, the movement or the post could also show hostility towards the counter companies.
\item \textit{Unknown} posts indicate that there is no clear indication, in either direction, that the comment supports the GameStop narrative. This can also be a catchall if the observation does not meet other criteria, such as unrelated posts.
\item An \textit{Informative} post is meant to inform users without any personal opinions or biases visible.  No emotions, no sides taken, only sharing information appears in the post.
\item The \textit{No} category indicates that the individual does not support GameStop.
\end{itemize}

An individual post can be a combination of any of these two categories. The following comment illustrates one combination found in our dataset:
\begin{center}
\noindent\fbox{%
    \parbox{.93\linewidth}{%
        ``I agree, I only one .02 shares of GME, and I did that specifically for this reason. After the squeeze, HF know what to expect going forward, so the momentum here is pretty much done. I'll hold my measly little .02 because I think once COVID ends we could see potential for growth, but I feel bad for those here that put thousands into this stock after the price was over \$200.''
    }
}
\end{center}
In this comment, the individual states they own some shares of GameStop outright, causing intent to be ``Yes.'' They also depict a lack of belief in the future success of GameStop and only hope. Showing a lack of support or belief that GME will continue its rise; therefore, support is rated as ``No.'' 

The following comment clearly supports the GME movement as he claims GME will moon (a term to suggest the stock will skyrocket); hence, support is ``Yes.'' 
\begin{center}
\noindent\fbox{%
    \parbox{.93\linewidth}{%
        ``You will make 10\% while watching everyone else moon on GME.''
    }
}
\end{center}
However, we cannot confirm one way or another, whether the individual has GME stock, so the Intent would be classified as ``Unknown.''

\begin{table}[t]
\centering
\resizebox{0.55\linewidth}{!}{%
\begin{tabular}{lrr}
\toprule
            & \textbf{Intent} & \textbf{Support} \\ \midrule
\textbf{Yes}         & 983    & 2473    \\
\textbf{No}          & 83     & 292     \\
\textbf{Maybe}       & 370    & N/A     \\
\textbf{Informative} & 318    & 257     \\
\textbf{Unknown}     & 3246   & 1973    \\ \cmidrule(lr){2-3}
\textbf{Total}       & 5000   & 5000    \\ \bottomrule
\end{tabular}%
}
\caption{Final Dataset Statistics}
\label{tab:data-stats}
\vspace{-1em}
\end{table}

After the annotation rules were agreed upon and the annotation guidelines were completed, annotation began on the set of 5000 randomly selected comments. Annotators completed the process independently of one another. Annotators completed 3000 annotations with an agreement. It was measured by Cohen's Kappa, of .81 for intent and .72 for support. Following the annotation, both annotators worked together to adjudicate comments that had differing annotations.
The remaining 2000 comments were then done without comparing to measure agreement by a single annotator. Finally, an outside third-party annotator was given the finalized instructions and asked to annotate a random set of 100 observations along with the annotation guidelines to measure the external validity of the annotation process. The third-party annotator achieved a Cohen's kappa of .76 for Intent and .65 for Support when compared to the 100 adjudicated annotations. Overall, there were a total of 5000 comments annotated, 3000 were annotated with two individuals, and 100 were annotated with three individuals.

The final dataset statistics can be found in Table~\ref{tab:data-stats}. We find that the majority of comments have an Intent of ``Unknown''. However, a large proportion of comments mention ``Yes'', while the ``No'' is the smallest Intent category. Intuitively, many comments are just discussing the event without discussing an actual purchase. But still, nearly one-fifth of the comments to express an intent to buy. We make similar findings for Support. But, ``Yes'' is the largest category, instead of ``Unknown.''. Again, this makes sense because the r/WallStreetBets started the GameStop hype. Hence, most of the members support it.

\section{Data Modeling}

In this section, we describe and evaluate several baseline models. Overall, our goal is to show that models can be trained to learn the categories we annotated. If models can not learn anything better than random, the dataset will be of little use to both social scientists and computational social science researchers.

\subsection{Baseline Models and Evaluation Metrics}

We explore four baselines on our dataset: a Linear SVM, RoBERTa, and two random baselines (Uniform and Stratified). We describe each baseline below:

\textbf{Linear SVM.} We trained a Linear SVM using the term frequency-inverse document frequency-weighting (TF-IDF) of unigrams and bigrams (i.e., single words, ``wsb'', and pairs of words like ``GameStop sucks'' are used as features) and L2 regularization. TF-IDF is a statistical measure that weights how important words are in a corpus. Furthermore, we searched for the best C value from the set \{0.0001, 0.001, 0.01, 0.1, 1, 10\} using a validation dataset. The SVM is implemented using the LinearSVC classifier in scikit-learn~\cite{pedregosa2011scikit}.

\textbf{RoBERTa.}  We fine-tuned RoBERTa~\cite{liu2019roberta} from the Huggingface libary~\cite{wolf2019huggingface}, specifically the roberta-base variant. Moreover, we used the last layer's CLS token which is passed to a softmax layer that is fine-tuned for up to 25 epochs. The model was checkpointed after each epoch, and the best version was chosen using the validation data. We used cross entropy loss as the objective function, a mini-batch size of 8, and learning rate of 2e-5 (other hyper-parameters same as \cite{liu2019roberta}). Finally, we used the Adam optimizer~\cite{kingma2014adam} with a Cosine linear learning rate scheduler~\cite{gotmare2018closer} with no warm-up steps.

\begin{table}[t]
\centering
\resizebox{.9\linewidth}{!}{%
\begin{tabular}{llrrr}
\toprule
                            &             & \textbf{Precision} & \textbf{Recall} & \textbf{F1} \\ \midrule
\textbf{Uniform Baseline} & \textbf{Macro AVG} & .197 &    .166    & .146   \\  
\textbf{Stratified Baseline} & \textbf{Macro AVG} &     .205 & .206   &  .205  \\ \midrule 
\multirow{6}{*}{\textbf{Linear SVM}} & Informative &     .744      &      .561  &   .640 \\
                            & Maybe       &    1.00       &   .054     &  .102  \\
                            & No          &      .000       & .000 &  .000  \\
                            & Unknown     &    .795       &       .912 &  .850  \\
                            & Yes         & .656          &       .660 &  .658  \\ \cmidrule(lr){2-5}
                               & \textbf{Macro AVG} & .639
                                      &   .437     &  .450  \\ \midrule 
\multirow{6}{*}{\textbf{RoBERTa}}    & Informative &      .714     &   .702     &  .708  \\
                            & Maybe       &      .329     &  .324      &  .326  \\
                            & No          &    .375       &     .187   &  .250  \\
                            & Unknown     &     .888      &    .893    &  .891  \\
                            & Yes         &     .793      &   .817     &  .805  \\ \cmidrule(lr){2-5}
                               & \textbf{Macro AVG}    & \textbf{.620}  &   \textbf{.585}     &  \textbf{.596}  \\ \bottomrule 
\end{tabular}%
}
\caption{Intent Results.}
\label{tab:intent-results}
\vspace{-1em}
\end{table}

\textbf{Random Baselines.} We use two random baselines from the scikit-learn package~\cite{pedregosa2011scikit}: Uniform and Stratified. The Uniform baseline makes predictions for each class with equal proportions. The Stratified random baseline makes predictions based on the class proportions in the training dataset.

\subsection{Experimental Setup}

We use a 70/10/20 split of the 5000 comments into a training, validation, and test dataset, respectively. Furthermore, we evaluate the model using precision, recall, and F1 Score for each class independently, along with the aggregate measure Macro F1-score.

\subsection{Results}

The results for \textit{Intent} are shown in Table~\ref{tab:intent-results}. Overall, we find that both the Linear SVM and RoBERTa outperform the random baselines with regards to the Macro F1 metric. Furthermore, RoBERTa outperforms the Linear SVM with regards to Macro F1 score by nearly 15\%. 
Unsurprisingly Unknown represents the easiest to predict as it was the largest class for our analysis. However, we are pleased to see a considerable increase in F1 score compared to the baseline, particularly for Yes, No, and Maybe classes. These Intent results show that machine learning models from our annotated data make the classes learnable. Hence, other researchers can use the predictions for potential downstream studies.

\begin{table}[t]
\centering
\resizebox{.9\linewidth}{!}{%
\begin{tabular}{llrrr}
\toprule
                            &             & \textbf{Precision} & \textbf{Recall} & \textbf{F1} \\ \midrule
\textbf{Uniform Baseline} & \textbf{Macro AVG} &   .251 & .196     &  .179 \\  
\textbf{Stratified Baseline} & \textbf{Macro AVG} &    .247 &  .247    &  .247  \\ \midrule 
\multirow{6}{*}{\textbf{Linear SVM}} & Informative &    .742       &  .479      &   .582 \\
                            & No          &     .250      &   .018     &  .033  \\
                            & Unknown     &      .591     &      .661  &  .624  \\
                            & Yes         &  .700         &     .732   &   .716 \\ \cmidrule(lr){2-5}
                               & \textbf{Macro AVG} & .571
                                      &    .473    &   .489 \\ \midrule 
\multirow{6}{*}{\textbf{RoBERTa}}    & Informative &      .808     &    .792    &  .800  \\
                            & No          &       .423    &    .196    &  .268  \\
                            & Unknown     &     .650  &     .717   &  .682  \\
                            & Yes         &       .777    &   .765     &  .771  \\ \cmidrule(lr){2-5}
                               & \textbf{Macro AVG}   & \textbf{.669}   &     \textbf{.617 }  & \textbf{.630}    \\ \bottomrule 
\end{tabular}%
}
\caption{Support Results.}
\label{tab:support-results}
\vspace{-1em}
\end{table}

For \textit{Support} results are shown in Table~\ref{tab:support-results}. Again, we find that both the Linear SVM and RoBERTa models outperform the random baselines, indicating that the classes are learnable. Specifically, we find that RoBERTa substantially outperforms the Linear SVM baseline by more than 14\% with regard to the Macro F1 score. Unsurprisingly, the worse performance is found for the ``No'' class with an F1 of .268 because it is the most infrequent within the dataset. Moreover, interestingly, the Informative class is the most accurate class for Support with an F1 of .800, followed by the most common class Yes with an F1 score of~.771.

\begin{table*}[t]
\centering
\resizebox{0.8\textwidth}{!}{%
\begin{tabular}{ll}
\toprule
       \textbf{Category}     & \textbf{Words} \\ \midrule 
\multicolumn{2}{c}{\textbf{Intent}} \\ \midrule
Yes         & my, bought, holding, me, my gme, im, to buy, i, buying, buying more                                       \\
Maybe       & we, i new, all in, time to, we are, us, gimme, 55k million, more tendies                                  \\
No          & puts, bag, gme bag, bag holders, holders, holders, don be, gme 12, bagholders are, gme bag holders, spite \\
Informative & https, com, robinhood, https www, www, cashapp, revolut, fidelity, allowing, on cashapp  \\ 
Unknown     & game, gme, you, their, they, everyone, more shares, he, gamestop, guy                                     \\ \midrule \midrule
       \multicolumn{2}{c}{\textbf{Support}} \\ \midrule
Yes         & hold,  \emojione, stop, we, fuck, moon, the moon, keep, love, 69                                                  \\
Informative & https, com, robinhood, on cashapp, cashapp, www, https www, fidelity, amc halted, was shorted             \\
Unknown     & game, what, was, why, does, gme thread, the same, then, isn, 50                                           \\
No          & bagholders, bag, puts, no gme, gme bagholders, cult, holders, is dead, sell amc, bag holders              \\
 \bottomrule
\end{tabular}%
}
\caption{Most predictive words found by the Linear SVM model for each Intent and Support category. The words are ranked based on predictive power, e.g., the first word is the most predictive, the second word is the second most predictive, etc.}
\label{tab:phrase}
\vspace{-1em}
\end{table*}

\subsection{Phrase Analysis}

In this section, we aim to find the most predictive phrases for each category based on the learned coefficents of the Linear SVM model. The results can be found in Table~\ref{tab:phrase}. There are a number of noteworthy and intuitive patterns. For example, we find that predictive words for the ``Yes'' intent category include ``bought'' and ``holding'', hence, indicating direct information about buying GME shares. Likewise, the ``No'' support category includes predictive words such as ``bagholders'' and ``puts'' (i.e., short sells) \footnote{The term holding refers to owning GME stock. The term bagholder is an insult and refers to individuals who purchased a stock at a high price and the price dropped considerably leaving the individual "holding the bag". The term put refers to betting against a stock (Betting the stock will decrease in value)} indicating negative valence towards the GameStop company or anti-support against the general GameStop narrative to fight the establishment. The ``Informative'' category for both Intent and Support show words that indicate information such as linking to a website. Interesting, the ``Maybe'' Intent category contains words plural pronouns (e.g., ``we'' and ``us'') making it unclear whether someone is showing an intent for themselves or someone else. Overall, this simple phrase analysis provides further evidence of the quality of the data annotation process by providing intuitive insights into each category beyond the annotation guidelines. 

\section {Discussion}
The initial study findings are promising. First, the high Cohen's kappa score during the annotation phrase indicates enough differentiation in the language syntax that it can be possible for humans to extract both the intent and support levels of the text. Furthermore, despite the colorful language used, the annotators were able to identify and agree on the interpretation of community-specific language such as sarcasm or other narrative wordplays. Other researchers have also explored the power of narrative-based discussions on Reddit~\cite{antoniak2019narrative}, which depicts how emotions and narratives unfold through language use on social media platforms. Next, the findings suggest that modeling with machine learning algorithms can perform substantially better than random baselines performance, indicating machine learning models can learn that data.  This is important for both NLP researchers as well as computation social scientists. For example, NLP researchers can use the data to further develop better algorithms. Likewise, social scientists can use the predictions of a model trained on our dataset to answer social and behavior questions (e.g., related to persuasion).

\section{Use Cases}

In this section, we describe two future research avenues and use cases for our dataset: Detecting cases of persuasion on Reddit and understanding persuasion methods that can change user Intent.

\subsection{Detecting Cases of Persuasion on Reddit}
Recent research into persuasion literature has generally utilized the Elaboration Likelihood Model (ELM) as an explanation of how an individual can be persuaded to behave. The Elaboration Likelihood Model suggests that when an individual arrives at a decision, the decision will either be based on the message and logical reasoning (the central route) or based on cues related to the message (the peripheral route.). This model (theory) has been important for social scientist to better understand human behavior as it is related to persuasion. One study on petitions from Change.Org found that cognitive reasoning and moral judgments do not lead to effective campaigns. Instead, successful persuasive campaigns rely on emotionally charged language and enlightening information~\cite{chen2019multi}. Furthermore, in another case study on textual conversations of CEOs and businesses, individuals process information from the peripheral route of the ELM would do so based on one of four appeals channels (social, ethical, political, and ideological)~\cite{pignot2020affective}. However, much of the current research on human persuasion tend to rely on subject survey’s asking the individual if the individual were persuaded \cite{yi2019leveraging}, at the aggregate \cite{chen2019multi,wang2021mitigating}, or attempt to model receptiveness to phrasing \cite{pignot2020affective}. The literature lacks a definitive, easily identified indicator of detecting intention to action. As a potential use case of this dataset, researchers can leverage the ELM (particularly the peripheral route) in qualitatively interpreting how an individual is persuaded to change their mind from a unsure intention to buy GME shares (Unknown) or might (maybe) to a positive intention (Yes). Moreover, researchers can expand the data to include qualitative longitudinal data of users who are confirmed to have made a stock purchase becoming a real life case of detected persuasion.  Another potential direction for the research would be to annotate more details such as the degree of support level (strongly opposed, strongly agree), capturing the degree of changed opinion over time.

\subsection{Practical Applications of Modeling Intent}
As previously stated, in modeling intent, particularly in Management Information Systems literature, there is a reliance on reporting perceptions such as ``Are you convinced'' or ``Would you buy this''~\cite{yi2019leveraging,gerlach2019they,yin2020anger}. On the other hand, our dataset attempts to minimize this element by identifying characteristics of individuals who are persuaded to engage in an activity. While there have been plenty of research using text analysis to outside of surveys, they tend to look at a more passive context such as  past consumer sentiment~\cite{jiang2021investigating}, or the success of a past event in the aggregate~\cite{wang2021social}. An interesting use of this dataset would be to identify persuasion at the individual level among peers.  Identifying the conversation patterns would allow extended research into cyber community behaviors where it would not be as obvious when the jump between intent and behavior is made, for example, from talk of threatening cybercrime to taking actually action offline. Applying the linguistics characteristics associated with committed behavior into more practical circumstances.

\section{Ethical Statement}
The authors of this paper acknowledge reading and abiding by the AAAI code of conduct and ethical guidelines for this submission. We acknowledge the risk that as our paper observes the human behavior of persuasion and social manipulation, those with unsavory intent can abuse the research for their ends. Nevertheless, we posit that research on how to manipulate and socially engineer is already available. Social manipulation is already a problem. Fake news, for example, is the major challenge for our time, and by not studying how manipulation occurs, we cannot learn how vulnerable we are and how to defend against it effectively. This dataset aims to provide a freely available corpus for peer-to-peer conversations that can confirm user support for an campaign (e.g., supporting GameStop) and user intentions to make a purchase (e.g., GME shares). Our research does not use any individuals personal information, nor any identifiable info to put the individual at risk. All information was taken from publicly available data on Reddit.
\section{FAIR Requirements}

Our dataset adheres to FAIR principles    (Findable, Accessible, Interoperable, and Re-usable). The dataset is Findable and Accessible through Zenodo~\footnote{\url{https://zenodo.org/record/5851847\#.YeO_vhPML8E}}. Moreover, the data is licensed under the Creative Commons Attribution License (CC BY 4.0). Finally, the data is shared as a CSV file along with the annotation guidelines, which are shared as word documents. Thus, the data is reusable and inter-operable.

\section{Conclusion}
We have seen that online social media communities are increasingly turning to collaboration as a mechanism to coordinate activities beyond the boundaries of the said community. The case of WallStreetBets suggests the importance of social media communication and its resulting influence on online human communications. This dataset provides the unique capability to not only measure intent but to reveal a tangible result, uniquely bridging intent, and action together. We contribute to the field of finance and IS using NLP based tools that can drive a new path forward in research addressing online collaboration behavior on social media platforms and potentially cyber crime.

\bibliography{refs.bib}
\bibliographystyle{aaai}

\begin{appendices}

\section{Appendix: Annotation Guidelines}

In this section, we provide the annotation guidelines that were developed for this project. Specifically, Table~\ref{tab:ann-intent} provides the guidelines for annotating Intent, while Table~\ref{tab:ann-support} provides the guidelines for Support.

\begin{table*}[t]
\centering
\resizebox{\textwidth}{!}{%
\begin{tabular}{p{7cm}p{7cm}p{7cm}p{7cm}}
\toprule
                                 & & & \\
\multicolumn{4}{p{28cm}}{For this process, we define intent as the   commentator exhibiting some interest in acquiring the Game Stop stock. For this   section intent can be already owning the stock, intent to purchase, or the   desire to not purchase the stock.} \\
\\
\multicolumn{4}{p{28cm}}{For example: ``Looks like my   puts on GME and AMC have me +14k today. Sorry! At some point GME and AMC   would go down. Hope no one is stuck holding the bag}\\
\\
\multicolumn{4}{p{28cm}}{In this example, the commentor is referring to   Game Stop stock holders as bag holders, a term used to describe the leftover   people who now have worthless stock. This strongly suggests the author has no   interest in the stock and would not purchase it. Furthermore, puts are a   method to bet against Game Stop price. So the desire to not purchase the stock is visible.} \\
\multicolumn{4}{p{7cm}}{} \\
\textbf{Annotation} & \textbf{Rules} & \textbf{Example(s)} & \textbf{Notes} \\ \midrule
\multirow{6}{*}{\textbf{Y = Yes}}          & \multicolumn{1}{p{7cm}}{There is clear intent to purchase or has already purchased GameStop stock in the recent time period.} & \multicolumn{1}{p{7cm}}{I'll send you my worth after it went 10x thanks   to the GME squeeze} & \multicolumn{1}{p{7cm}}{The individual indicates his worth will increase,   so it is clear they currently own GameStop shares} \\
                                 & & & \\
                                 & & \multicolumn{1}{p{7cm}}{Fuck it, my tax return is going back to GME.} & \multicolumn{1}{p{7cm}}{In this example, the individual outright states   his intent to use his tax return to invest in game stop in the future.} \\
                                 & & & \\
                                 & & \multicolumn{1}{p{7cm}}{Where can i still buy GME Shares?} & \multicolumn{1}{p{7cm}}{Again the author intends to buy GameStop but does not know where.} \\
                                 & & & \\ \midrule
\multirow{6}{*}{\textbf{M = Maybe}}         & \multicolumn{1}{p{7cm}}{It is not clear the individual has the stock, but the context hints of a possibility of purchase or already owns the stock}                                                                                                & \multicolumn{1}{p{7cm}}{Come listen to the GME WAR ROOM RADIO, for market   open !!! Hold the line apes only https://www.twitch.tv/stashkonig} & \multicolumn{1}{p{7cm}}{``hold the line'' implying he could be someone who has the stock, but it can not be determined with certainty.''} \\
                                 & & & \\
                                 & & \multicolumn{1}{p{7cm}}{CNBC try hards trying shitting on   Reddit and GME so hard, let's shit on them, BUY OR HOLD, there is no sell} & \multicolumn{1}{p{7cm}}{In this comment, its not entirely certain if the individual owns or will purchase GME but its clear he leans towards the buy side.} \\
                                 & & & \\
                                 & & \multicolumn{1}{p{7cm}}{Worked for me yesterday. Was up and funded instantly. GME appears to be buyable} & \multicolumn{1}{p{7cm}}{This comment, the author suggests GME can be bought and that he may have done it yesterday; but it could also be that the app worked. So it leans towards buying intent.} \\
                                 & & & \\ \midrule
\multirow{6}{*}{\textbf{U = Unknown}}      & \multicolumn{1}{p{7cm}}{It is not clear one way or another the intent to   purchase or currently owning the stock. This can serve as a catchall if   unable to annotate to any other category, such as completely unrelated posts.}                  & \multicolumn{1}{p{7cm}}{ITS ALL ABOUT GME!!!! BUY AND HOLD} & \multicolumn{1}{p{7cm}}{This clearly supports GME, but there is no   indication that the person might already possess the stock} \\
                                 & & & \\
                                 & & \multicolumn{1}{p{7cm}}{Google gamestop news today} & \multicolumn{1}{p{7cm}}{This post does not reveal much information, nor does it inform.} \\
                                 & & & \\
                                 & & \multicolumn{1}{p{7cm}}{yo l can finally find some DD on   stocks not named gme thank the fucking god} & \multicolumn{1}{p{7cm}}{This post somewhat hints in the lack of   disinterest in GameStop, however it is not informative enough to draw a   conclusion.} \\
                                 & & & \\ \midrule
\multirow{6}{*}{\textbf{I = Informative}}  & \multicolumn{1}{p{7cm}}{Post is meant to inform users without any   personal opinions or biases visible. No emotions, no sides taken, only sharing information}                                                                                   & \multicolumn{1}{p{7cm}}{GME halted    10:13:47} & \multicolumn{1}{p{7cm}}{This post only informs when GME stock had stopped   trading. It is simple and informative.} \\
                                 & & & \\
                                 & & \multicolumn{1}{p{7cm}}{Probably not the most popular, but   I bookmark the on Yahoo Finance. I like that it shows   pre-market. I've not done much hunting for anything else.} & \multicolumn{1}{p{7cm}}{This example only informs users of a site to look   up stocks. The intent is clearly to help users utilize a resource. There is   no side taken for GME intent.} \\
                                 & & & \\
                                 & & \multicolumn{1}{p{7cm}}{**BZ: GameStop Shares Expected To Resume Trade From Circuit Breaker At   12:46:02 p.m. EST**} & \multicolumn{1}{p{7cm}}{This post is acting as second hand reporting.} \\
                                 & & & \\ \midrule
\multirow{6}{*}{\textbf{N = No}}           & \multicolumn{1}{p{7cm}}{The post indicates clear disinterest, or no   intention to purchase the stock.  The   individual does not and will not purchase the stock.} & \multicolumn{1}{p{7cm}}{All it is, is a case of Misery loves company its   why GME bag holders still hyping it up and hating on everyone taking interest   in other stocks lol} & \multicolumn{1}{p{7cm}}{This individual refers to people holding GME as   bagholders (someone who will lose money) indicating that he does not own the   stock and considers it a scheme.} \\
                                 & & & \\
                                 & & \multicolumn{1}{p{7cm}}{Lol yall still doing this gme   thing huh? We had the greenest week in 8 months and you think your money   should be invested in a pump and dump? Man, I almost feel bad for you, then I   remember that you had 30 warnings and chose to ignore them. Have fun!} & \multicolumn{1}{p{7cm}}{In this case, the writer clearly considers   GameStop a scam and has no interest in holding.} \\
                                 & & & \\
                                 & & \multicolumn{1}{p{7cm}}{Tesla in some weird way could   actually be the company of the future. They certainly aren't right   now but it's not out of the realm of possibility. GameStop has no future,   even if they pivoted entirely to digital sales and e-commerce, nothing could   possibly justify a \$20B valuation, with the information they have currently   released to the public.}                                                                & \multicolumn{1}{p{7cm}}{While not outright said, this individual suggests   Gamestop is a dying company and vastly overvalued; thus it can be strongly   inferred the individual has no interest in the stock.} \\
                                 & & & \\ \bottomrule
\end{tabular}%
}
\caption{Annotation Guidelines for \textit{Intent}}
\label{tab:ann-intent}
\end{table*}

\begin{table*}[t]
\centering
\resizebox{\textwidth}{!}{%
\begin{tabular}{p{7cm}p{7cm}p{7cm}p{7cm}}
\toprule
                                 & & & \\
\multicolumn{4}{p{28cm}}{For this process, support is defined as any indication that the commentor has some support for the Game Stop company. Support can encompass either supporting the Game Stop company itself, that is, believing in the worth of the firm, support for the stock to increase, using supportive terminology (such as ``to the moon'' or ``GME GME GME'' cheers); or it can include degree of support for the narrative of ``us vs. them''. } \\
\\
\multicolumn{4}{p{28cm}}{For example: They're thinking: launch invesitigations and take every reddit wsb user to court that got GME for the sole purpose of forcing you to sell your GME to cover legal fees, therefore dropping the stock price so the shorts can benefit.  HAHA They crazy if they think this group won't band together. I see strength here Ihave never seen on ws. Stay strong!}\\
\\
\multicolumn{4}{p{28cm}}{In this example the individual first implies that courts are against GME leading to the belief of support; however, the comment goes on to suggest themes such as banding together and stay strong. These are direct indicators of support for the ``Us vs Them'' theme.} \\
\multicolumn{4}{p{7cm}}{} \\
\textbf{Annotation} & \textbf{Rules} & \textbf{Example(s)} & \textbf{Notes} \\ \midrule
\multirow{6}{*}{\textbf{Y = Yes}} & \multicolumn{1}{p{7cm}}{The post indicates clear support for the GME narrative and for GameStop as a company.
The post could also show hostility towards the counter companies.
meStop stock in the recent time period.} & \multicolumn{1}{p{7cm}}{GME was also predicted to hit 1k or more. } & \multicolumn{1}{p{7cm}}{In example one, the post shows clear support for GME, believing the stock will rise. } \\
                                 & & & \\
                                 & & \multicolumn{1}{p{7cm}}{i agree GME will stay at the same levels as it is now for long period of time, but there is still probability it will jump very high!
Hedge funds will not scare us! GME, AMC, NOK, BB!} & \multicolumn{1}{p{7cm}}{In example two, The support is indicated through antagonism against the hedge funds (the companies betting against GameStop), promoting the ``Us vs Them'' narrative. This sentiment is positive} \\
                                 & & & \\
                                 & & \multicolumn{1}{p{7cm}}{CNBC try hards trying shitting on Reddit and GME so hard, let's shit on them, BUY OR HOLD, there is no sell.} & \multicolumn{1}{p{7cm}}{Example 3 suggests an ``us vs them'' against the general media supporting the GameStop narrative.} \\
                                 & & & \\ \midrule
\multirow{6}{*}{\textbf{U = Unknown}}      & \multicolumn{1}{p{7cm}}{There is no clear indication, in either direction, that the comment supports the Gamestop narrative. This can also be a catchall if the observation does not meet other criteria, such as completely unrelated posts..}                  & \multicolumn{1}{p{7cm}}{if I got it right GME has 122\% of float shorted, AMC has 27\% ... it's why I'm asking someone to confirm} & \multicolumn{1}{p{7cm}}{This comment does not indicate support. It attempts to inform but gives the individuals opinion/calculations. Therefore it is not an informative post either} \\
                                 & & & \\
                                 & & \multicolumn{1}{p{7cm}}{\$GME isn't available on cashapp afaik.} & \multicolumn{1}{p{7cm}}{This comment leaves no indication on support at all. It makes a claim and could be informative but provides uncertainty so it does not qualify.} \\
                                 & & & \\
                                 & & \multicolumn{1}{p{7cm}}{Gonna assume it will get deleted. AMC crack down started now GME crack downs.} & \multicolumn{1}{p{7cm}}{This comment is only commenting on their opinion of events, it does not imply any support.} \\
                                 & & & \\ \midrule
\multirow{6}{*}{\textbf{I = Informative}}  & \multicolumn{1}{p{7cm}}{Post is meant to inform users without any personal opinions or biases visible.  No emotions, no sides taken, only sharing information.}                                                                                   & \multicolumn{1}{p{7cm}}{Probably not the most popular, but I bookmark the GME on Yahoo Finance. I like that it shows pre-market. I've not done much hunting for anything else.
} & \multicolumn{1}{p{7cm}}{This post recommends a site to analyze GME data. It does not indicate any bias or interest, simply a source recommendation.} \\
                                 & & & \\
                                 & & \multicolumn{1}{p{7cm}}{If you've got a Revolut account you can get them on their, and GME} & \multicolumn{1}{p{7cm}}{This post recommends an app to purchase GME, it does not indicate any bias or interest.} \\
                                 & & & \\
                                 & & \multicolumn{1}{p{7cm}}{Easier to calculate the overall breakeven using the total in/out

He spent 90k overall on all the contracts combined, so he needs at least 90k out to breakeven.

He has 10 each of 275, 285. For each dollar above each strike he can get \$1000 back, so he needs to be a total of \$90 in the money across all the strikes. 90 = (x - 275) + (x - 285) = 2x - 560

x = (90+560)/2 = 325, so he needs GME at \$325 at expiry to breakeven
} & \multicolumn{1}{p{7cm}}{This post is focused on how a specific financial vehicle can be calculated. The post is meant to inform and does not indicate any support.} \\
                                 & & & \\ \midrule
\multirow{6}{*}{\textbf{N = No}}           & \multicolumn{1}{p{7cm}}{The individual does not support Game Stop.} & \multicolumn{1}{p{7cm}}{Amc and nok, gme is broken} & \multicolumn{1}{p{7cm}}{The first post shows a lack of support for GameStop independent of intent.} \\
                                 & & & \\
                                 & & \multicolumn{1}{p{7cm}}{I'm holding because i wouldn't gain much from selling at this point,  but raw number comparisons to VW were always bad logic. sure, VW hit 1k, but it strated at 250. Gamestop rocketing from 40 to almost 400 was already a much larger jump. Will it go up again, idk. it might, but I think some here are too confident in that.  But if we are waiting for the moment's, then we barreled right past it last week.} & \multicolumn{1}{p{7cm}}{The second post, clearly suggests owning GameStop stock, but the individual does not seem to support the company, or the stock increasing.} \\
                                 & & & \\
                                 & & \multicolumn{1}{p{7cm}}{Idk, probably going to be called a bot, my guess GME has been squeezed already. I'm still holding on for another but I think its probably just starting its slow descent back down. I'm not very knowledgeable in the stock market's more complicated areas. Just super basic stuff.  Still learning more everyday so feel free to correct me this is not financial advice yada yada yada.}                                                                & \multicolumn{1}{p{7cm}}{This post is interesting as its clear the individual owns the GameStop stock but he does not believe in the company. Instead the individual believes the stock will decrease strongly implying that he thinks the stock increase was just an event and not the actual value of the company.} \\
                                 & & & \\ \bottomrule
\end{tabular}%
}
\caption{Annotation Guidelines for \textit{Support}}
\label{tab:ann-support}

\end{table*}

\end{appendices}

\end{document}